\documentclass[11pt,letterpaper]{article}
\usepackage{emnlp2017}
\usepackage{times}
\usepackage{latexsym}
\usepackage{amsmath}
\usepackage{graphicx}
\usepackage{bm}
\usepackage{amssymb}
\usepackage{footnote}
\usepackage{hyperref}

\emnlpfinalcopy

\graphicspath{{./}}

\def\emnlppaperid{***}

\title{An Embedded Deep Learning based Word Prediction}

\author{ Seunghak Yu\Thanks{\enskip Equal contribution} \qquad  Nilesh Kulkarni\footnotemark[1] \qquad Haejun Lee \qquad Jihie Kim \\
	Samsung Electronics Co. Ltd., South Korea\\
      {\tt \{seunghak.yu, n93.kulkarni, haejun82.lee, jihie.kim\}@samsung.com}}
\date{}

\begin{document}

\maketitle

\begin{abstract}
Recent developments in deep learning with application to language modeling have led to success in tasks of text processing, summarizing and machine translation. 
However, deploying huge language models for mobile device such as on-device keyboards poses computation as a bottle-neck due to their puny computation capacities. 
In this work we propose an embedded deep learning based word prediction method that optimizes run-time memory and also provides a real time prediction environment. Our model size is 7.40MB and has average prediction time of 6.47 ms. We improve over the existing methods for word prediction in terms of key stroke savings and word prediction rate.
\end{abstract}

\section{Introduction}

Recurrent neural networks (RNNs) have delivered state of the art performance on language modeling (RNN-LM) \cite{mikolov2010recurrent,kim2015character,miyamoto2016gated}. 
A major advantage of RNN-LMs is that these models inherit the property of storing and accessing information over arbitrary context lengths from RNNs \cite{karpathy2015visualizing}.   
The model takes as input a textual context and generates a probability distribution over the words in the vocabulary for the next word in the text. 

However, the state of the art RNN-LM requires over 50MB of memory (\cite{zoph2016neural} contains 25M parameters; quantized to 2 bytes). This has prevented deploying of RNN-LM on mobile devices for word prediction, word completion, and error correction tasks. Even on high-end devices, keyboards have constraints on memory (10MB) and response time (10ms), hence we cannot apply RNN-LM directly without compression.

Various deep model compression methods have been developed. Compression through matrix factorization \cite{sainath2013low,xue2013restructuring,nakkiran2015compressing,prabhavalkar2016compression,lu2016learning} has shown promising results in model compression but has been applied to the tasks of automatic speech recognition. Network pruning \cite{lecun1989optimal,han2015deep,han2015learning} keeps the most the relevant parameters while removing the rest. Weight sharing \cite{gong2014compressing,chen2015compressing,ullrich2017soft} attempts to quantize the parameters into clusters. Network pruning and weight sharing methods only consider memory constraints while compressing the models. They achieve high compression rate but do not optimize test time computation and hence, none of them are suitable for our application.

\begin{figure*}[h]
  \includegraphics[width=\textwidth]{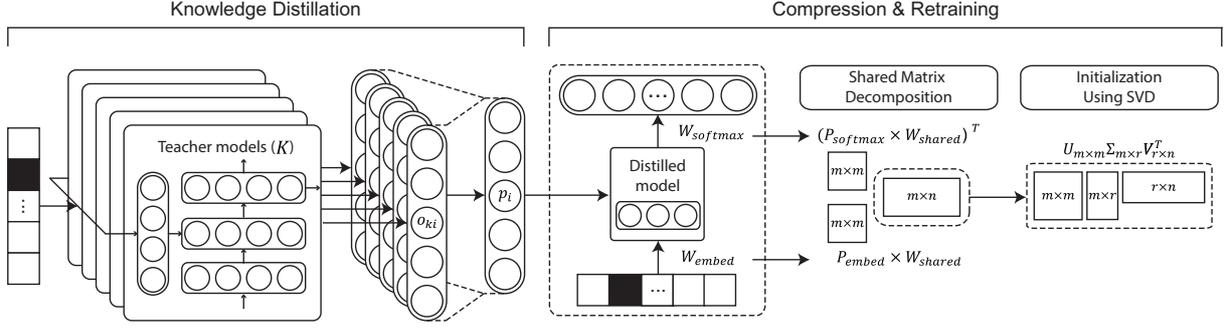}
  \centering
\caption{Overview of the proposed method. \bm{$o_{ki}$}: the $i_{th}$ logits of $k_{th}$ model, \bm{$p_i$}: the $i_{th}$ softened output of ensemble. $(P_{softmax} \times W_{shared})^T$ and $P_{embed} \times W_{shared}$ substitute $W_{softmax}$ and $W_{embed}$ in the proposed model respectively.}
  \label{overview}
\end{figure*}

To address the constraints of both memory size and computation, we propose a word prediction method that optimizes for run-time, and memory to render a smooth performance on embedded devices. We propose shared matrix factorization to compress the model along with using knowledge distillation to compensate the loss in accuracy while compressing. The resulting model is approximately 8$\times$ compressed with negligible loss in accuracy and has a response time of 6.47ms per prediction. To the best of our knowledge, this is the first approach to use RNN-LMs for word prediction on mobile devices whereas previous approaches used n-gram based statistical language models \cite {klarlund2003word,tanaka2007word} or unpublished. We achieve better performance than existing approaches in terms of Key Stroke Savings (KSS) \cite{fowler2015effects} and Word Prediction Rate (WPR). The proposed method has been successfully commercialized. 

\section{Proposed Method}

\subsection {Overview}
Figure \ref{overview} shows an overview of our approach. We propose a pipeline to compress RNN-LM for on-device word prediction with negligible loss of accuracy. 
Following sections describe each steps of our method. In Section \ref{basic-lm}, we describe the basic architecture of language model which is used as an elementary model in our pipeline. In Section \ref{student-teacher}, we describe method to make a distilled model by knowledge distillation and compensate for loss in accuracy due to compression. Following Section \ref{matrix-decomposition} describes model compression strategies to reduce memory usage and run-time.

\subsection{Baseline Language Model} \label{basic-lm}
\begin{figure}[h]
  \includegraphics[width=\columnwidth]{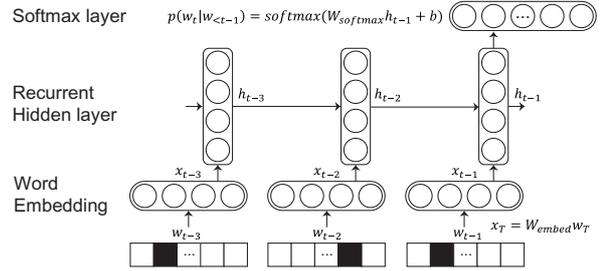}
\caption{Conventional RNN-LM.}
\label{model}
\end{figure}

All language models in our pipeline mimic the conventional RNN-LM architecture as in Figure \ref{model}. Each model consists of three parts: word embedding, recurrent hidden layers, and softmax layer. Word embedding \cite{mikolov2013distributed} takes input word $w_{t}$ at time $t$ as one hot vector and maps it to $x_{t}$ in continuous vector space $\mathbb{R}^{d}$. This process is parametrized by embedding matrix as $x_{t} = W_{embed}w_{t}$, where $W_{embed} \in \mathbb{R}^{d \times |V|}$, $V$ is the vocabulary, and $d$ is the dimension of embedding space.

The embedded word $x_{t}$ is input to LSTM based hidden layers. We use the architecture similar to the non-regularized LSTM model by \cite{zaremba2014recurrent}.
The hidden state of the LSTM unit $h_{t}$ is affine-transformed by the softmax function, which is a probability distribution over all the words in the $V$ as in Eq. \ref{softmax}.
\begin{equation}\label{softmax}
p(w_{t} = i | w_{< t}) = \frac{\exp(W_{i}^{T}h_{t-1}+ b_{i})}{\sum_{j} \exp(W_{j}^{T}h_{t-1}+ b_{j})}
\end{equation}

We train the model with cross-entropy loss function using Adam \cite{kingma2014adam} optimizer. Initial learning rate is set to 0.001 and decays with roll-back after every epoch with no decrement in perplexity on the validation dataset.

\subsection{Distilling Language Model} \label{student-teacher}
Knowledge Distillation (KD) \cite{hinton2015distilling} uses an ensemble of pre-trained teacher models (typically deep and large) to train a distilled model (typically shallower). KD helps provide global information to distilled model, and hence regularizes and gives faster updates for the parameters.

We refer to `hard targets' as true labels from the data. Contrary to the baseline model which only uses `hard targets', we adapt KD to learn a combined cost function from `hard targets'  and `soft targets'.  `Soft targets' are generated by adding a temperature $T$ (Eq.\ref{dark_knowledge}) to averaged logits of teachers' $z_i$ to train distilled model.
\begin{equation}\label{dark_knowledge}
p_i = \frac{\exp(\frac{z_i}{T})}{\sum_{j} \exp(\frac{z_j}{T})} \quad (when \; z_i = \overline{o_{ki}})
\end{equation}

Experiments in Table \ref{pipeline_experiment} shows improvement in perplexity compared to the models trained only with `hard targets'. We also use the combined cost function to retrain the model after compression. Retraining with combined cost function compensates for the loss in performance due to compression proposed in Section \ref{matrix-decomposition}.

\begin{table}[t]
\begin{center}
\begin{tabular}{ cccc }
\hline
\hline
 Language & Source & Words & Sentences\\
 \hline
 EN & Reddit & 1.1B & 71.2M\\
 EN & Twitter & 0.9B & 55.2M\\
 \hline
\end{tabular}
\end{center}
\caption{Collected data for language modeling.}
\label{dataset}
\end{table}

\subsection{Shared Matrix Factorization} \label{matrix-decomposition}

We present a compression method using shared matrix factorization for embedding and softmax layers in a RNN-LM. In the language model word embedding is trained to map words with similar context into a solution space closely, while softmax layer maps context to similar words. Therefore, we assume we can find sharable parameters that have characteristics similar to both embedding and softmax. Recently, there have been preprints \cite{press2016using,inan2016tying} suggesting an overlap of characteristics between embedding and softmax weights. 

We facilitate sharing by $W_{shared}$ across softmax and embedding layers, allowing for more efficient parameterization of weight matrices. This reduces the total parameters in embedding and softmax layers by half. We introduce two trainable matrices $P_{embed}$ and $P_{softmax}$, called the projection matrices, that adapt the $W_{shared}$ for the individual tasks of embedding and softmax as in Eq. \ref{shared}.

\begin{equation}
\begin{aligned}\label{shared}
W_{embed} &= P_{embed} W_{shared}\\
W_{softmax} &= (P_{softmax} W_{shared})^{T}
\end{aligned}
\end{equation}

%Furthermore, in the layers parametrized by $W_{shared}$ only a few outputs are active for a given input, we suspect that they are probably correlated and the underlying weight matrix has low rank $r$. There exists a factorization of $W_{m \times n} = W^{A}_{m \times r}W^{B}_{r \times n}$  where $W^{A}$ and $W^{B}$ are full rank \cite{strang1993introduction}. In our low-rank compression strategy, we expect rank of $W$ as $r'$ and compute the compression rate $c$ as $(mr' + r'n) / mn$. This compression scheme, without loss of generality is applied to $W_{shared}$.

Furthermore, in the layers parametrized by $W_{shared}$ only a few outputs are active for a given input, we suspect that they are probably correlated and the underlying weight matrix has low rank $r$. For such a weight matrix, $W$, there exists a factorization of $W_{m \times n} = W^{A}_{m \times r}W^{B}_{r \times n}$  where $W^{A}$ and $W^{B}$ are full rank \cite{strang1993introduction}. In our low-rank compression strategy, we expect rank of $W$ as $r'$ which leads to factorization in Eq. \ref{factorization}.

\begin{equation}\label{factorization}
W_{m \times n} \approx W^{A}_{m \times r'}W^{B}_{r' \times n}
\end{equation}

%Approximation in Eq. \ref{factorization} during factorization leads to degradation in model accuracy but when followed by fine-tuning through retraining it results in restoration of accuracy. Moreover, we compress by applying Singular Value Decomposition (SVD) to $W_{shared}$ in the non-compressed model to initialize the decomposed matrices for compressed model. SVD has been proposed as a promising method to perform factorization for low rank matrices \cite{nakkiran2015compressing,prabhavalkar2016compression}. We apply SVD on $W_{m \times n}$ to decompose it as $W_{m \times n} = U_{m \times m} \Sigma_{m \times n} V_{n \times n}^{T}$. $U, \Sigma, V$ are used to initialize $W^{A}$ and $W^{B}$ for the retraining process. We use the top $r'$ singular values from $\Sigma$ and corresponding $r'$ rows from $V^{T}$. Therefore, $W^{A} = U_{m \times m} \Sigma_{m \times r'} $ and $W^{B} =  V^{T}_{r' \times n}$, we  replace all the linear transformations using $W_{m \times n}$ with $W^{A} \times W^{B}$. 

Moreover, we compress by applying Singular Value Decomposition (SVD) to initialize the decomposed matrices. SVD has been proposed as a promising method to perform factorization for low rank matrices \cite{nakkiran2015compressing,prabhavalkar2016compression}. We apply SVD on $W_{m \times n}$ to decompose it as $W_{m \times n} = U_{m \times m} \Sigma_{m \times n} V_{n \times n}^{T}$. $U, \Sigma, V$ are used to initialize $W^{A}$ and $W^{B}$ for the retraining process. We use the top $r'$ singular values from $\Sigma$ and corresponding $r'$ rows from $V^{T}$. Therefore, $W^{A} = U_{m \times m} \Sigma_{m \times r'} $ and $W^{B} =  V^{T}_{r' \times n}$, we replace all the linear transformations using $W_{m \times n}$ with $W^{A} \times W^{B}$. Approximation in Eq. \ref{factorization} during factorization leads to degradation in model performance but when followed by fine-tuning through retraining it results in restoration of accuracy. This compression scheme, without loss of generality is applied to $W_{shared}$.

\section{Experiment Results}
\subsection{Evaluation of proposed approach}
Table \ref{dataset} describes the source of dataset\footnote{The dataset is available at \url{https://github.com/meinwerk/WordPrediction}}, number of words and sentences. This data is extracted from resources on the Internet, in a raw form with 8 billion words. We uniformly sample 10\% (196 million) from the dataset. It consists of 60\% for training, 10\% for validation and 30\% for test.

We preprocess raw data to remove noise and filter phrases. We also replace numbers in the dataset with a special symbol, $\langle NUM\rangle$ and out-of-vocabulary (OOV) words with $\langle UNK\rangle$. We append start of sentence token $\langle s\rangle$ and end of sentence token $\langle/s\rangle$ to every sentence. We convert our dataset to lower-case to increase vocabulary coverage and use top 15K words as the vocabulary.

\begin{table}[t]
\begin{center}
\begin{tabular}{ lccc}
\hline
\hline
Model & PP & Size & CR\\
\hline
Baseline & 56.55 & 56.76 & - \\
+ KD & 55.76 & 56.76 & -\\
+ Shared Matrix & 55.07& 33.87& 1.68$\times$\\
+ Low-Rank, Retrain  & 59.78 &  14.80&  3.84$\times$\\
+ Quantization & $\sim${\bf 59.78}& {\bf 7.40} & {\bf 7.68$\times$}\\
 \hline
\end{tabular}
\end{center}
\caption{Evaluation of each model in our pipeline. Baseline uses `hard targets' and Knowledge Distillation (KD) uses `soft targets'. Size is in MB and 16 bit quantization is applied to the final model. PP: Word Perplexity, CR: Compression Rate.}
\label{pipeline_experiment}
\end{table}

Table \ref{pipeline_experiment} shows evaluation result of each step in our pipeline. We empirically select 600 embedding dimension, single hidden layer with 600 LSTM hidden units for baseline model. Word Perplexity is used to evaluate and compare our models. Perplexity over the test set is computed as $\exp(-\frac{1}{N}\sum_{i=1}^{N} \log p(w_{i}| w_{<i}))$, where N is the number of words in the test set. Our final model is roughly $8 \times$ smaller than the baseline with 5\% (3.16) loss in perplexity.

\subsection{Performance Comparison}  

We compare our performance with existing word prediction methods using manually curated dataset\footnote{The dataset consists of 102 sentences (926 words, 3,746 characters) which are collection of formal and informal utterances from various sources. It is available at \url{https://github.com/meinwerk/WordPrediction}}, which covers general keyboard scenarios. Due to lack of access to language modeling engine used in other solutions, we are unable to compare word perplexity. To the best of our efforts, we try to minimize all the personalization these solutions offer in their prediction engines. We performed human evaluation on the manually curated dataset. We employed three evaluators from the inspection group to cross-validate all the tests in Table \ref{KSS_experiments} to eliminate human errors.

\begin{table}[t]
\begin{center}
\begin{tabular}{ lcc }
\hline
\hline
Developer & KSS(\%) & WPR(\%)\\
\hline
Proposed  & {\bf 65.11} & {\bf 34.38}\\
Apple  & 64.35 & 33.73\\
Swiftkey & 62.39 & 31.14\\
Samsung  & 59.81 & 28.84\\
Google & 58.89 & 28.02\\
 \hline
\end{tabular}
\end{center}
\caption{Performance comparison of proposed method and other commercialized keyboard solutions by various developers.}
\label{KSS_experiments}
\end{table}

We achieve the best performance compared to other solutions in terms of Key Stroke Savings (KSS) and Word Prediction Rate (WPR) as shown in Table \ref{KSS_experiments}. KSS is a percentage of key strokes not pressed compared to a keyboard without any prediction or completion capabilities. Every character the user types using the predictions of the language model counts as key stroke saving. WPR is percentage of correct word predictions in the dataset.

\begin{figure}[t]
  \includegraphics[width=\columnwidth]{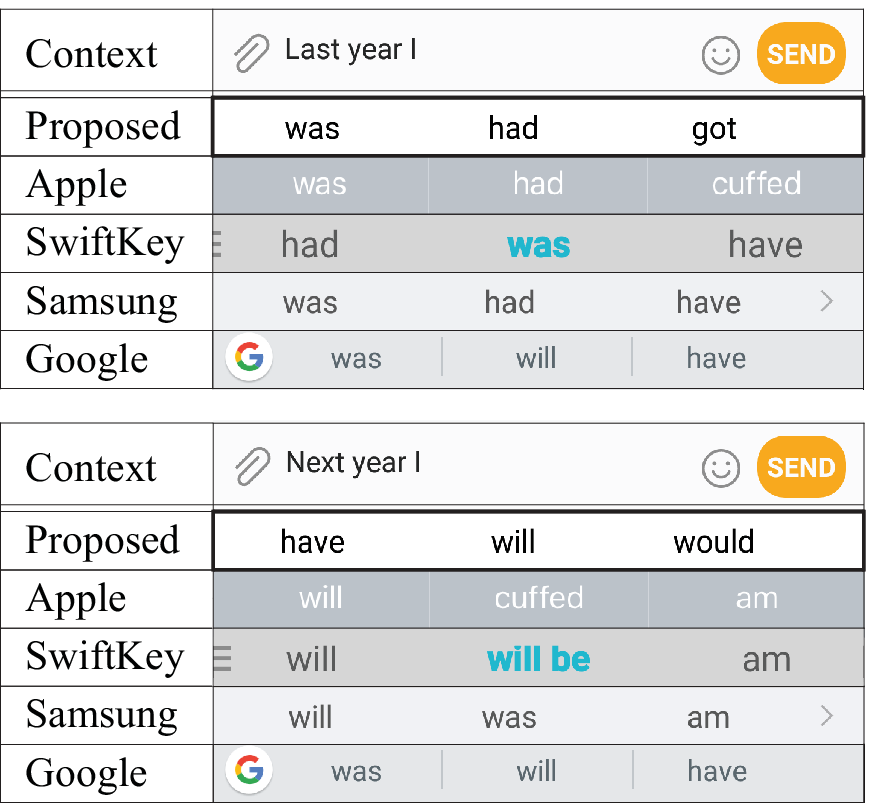}
  \caption{Example of comparision with other commercialized solutions. Predicted words for the contexts ``Last year I'' and ``Next year I''.}
\label{keyboard}
\end{figure}

While evaluating KSS and WPR, the number of predictions for the next word is same for all the solutions. The proposed method shows 65.11\% in terms of KSS and 34.38\% in WPR which is the best score among the compared solutions. For example, if the user intents to type ``published" and types only 34.89\% characters (``pub"), one of the top two predictions is ``published''. Furthermore, 34.38\% words the user intents to type are among the top three predictions. Figure \ref{keyboard} shows an example of word prediction across different solutions. In this example, we can spot some grammatical errors in the predictions from other solutions.

\section{Conclusions and Future Work}	
%We have proposed a practical method for training and deploying RNN-LM for mobile device which can satisfy memory and runtime constrains. Our method makes generalized distilled model by using averaged output of teachers and compresses its weight matrices by applying shared matrix factorization. We achieve 7.40MB in memory size and 6.47ms in average prediction time. Also, we have compared proposed method to existing commercialized keyboards in terms of key stroke savings and word prediction rate. In our benchmark tests, our method out-performed the others. 

We have proposed a practical method for training and deploying RNN-LM for mobile device which can satisfy memory and runtime constrains. Our method utilizes averaged output of teachers to train a distilled model and compresses its weight matrices by applying shared matrix factorization. We achieve 7.40MB in memory size and satisfy the run time constraint of 10ms in average prediction time (6.47ms). Also, we have compared proposed method to existing commercialized keyboards in terms of key stroke savings and word prediction rate. In our benchmark tests, our method out-performed the others. 

RNN-LM does not support personalization independently. However, our model which is currently commercialized uses RNN-LM along with n-gram statistics to learn user's input pattern and uni-gram to cover OOV words. Future work is required on directly personalizing the RNN-LM model to user's preferences rather than interpolating it with n-gram statistics to take full advantage.

\section*{Acknowledgments}
\bibliography{emnlp2017}
\bibliographystyle{emnlp_natbib}

\end{document}